\pdfoutput=1
\RequirePackage{xcolor}
\documentclass[journal,twoside,web]{ieeecolor}
\usepackage{tmi}
\usepackage{cite}

\usepackage{amsmath,amssymb,amsfonts}
\usepackage{algorithm}
\usepackage{graphicx}
\usepackage{textcomp}
\usepackage{soul}
\def\BibTeX{{\rm B\kern-.05em{\sc i\kern-.025em b}\kern-.08em
    T\kern-.1667em\lower.7ex\hbox{E}\kern-.125emX}}

\usepackage{url}
\usepackage{bbm}
\usepackage{ctable}
\usepackage{multirow}
\usepackage[font={footnotesize,sf}]{subcaption}
\usepackage[font={footnotesize,sf}]{caption}
\usepackage{cleveref}

\graphicspath{{figures/}}
\DeclareGraphicsExtensions{.pdf,.png}

\newcommand{\alignedintertext}[1]{%
  \noalign{%
    \vskip\belowdisplayshortskip
    \vtop{\hsize=\linewidth#1\par
    \expandafter}%
    \expandafter\prevdepth\the\prevdepth
  }%
}

\usepackage{graphicx}
\usepackage{xfrac,wrapfig,comment}
\usepackage{mathtools}

\usepackage{dsfont}
\usepackage{booktabs,paralist,hhline,colortbl}
\usepackage{algpseudocode}

\algnewcommand\algorithmicinput{\textbf{Input:}}
\algnewcommand\algorithmicoutput{\textbf{Output:}}
\algnewcommand\Input{\item[\algorithmicinput]}%
\algnewcommand\Output{\item[\algorithmicoutput]}%
\newcommand{\algrule}[1][.2pt]{\par\vskip.5\baselineskip\hrule height #1\par\vskip.5\baselineskip}

\usepackage{graphicx}
\usepackage{soul}

\newcommand{\mcc}[1]{\multicolumn{1}{c}{#1}} % short for multicolumn-centered
\newcommand{\etal}{\mbox{\emph{et al.}}}
\newcommand{\bt}[1]{\textbf{#1}}

\begin{document}
% \linenumbers
\bstctlcite{IEEEexample:BSTcontrol}
\title{Echocardiography Segmentation Using Neural ODE-based Diffeomorphic Registration Field}
\author{Phi V. Nguyen, Hieu H. Pham*, Long Q. Tran*
\thanks{Manuscript received November xx, 2021; revised April xx, 2022. This work was supported in part by the xx }
\thanks{*Correspondence authors}
\thanks{This work was supported by the Vingroup Innovation Foundation (VINIF) under project code VINIF.2019.DA02 and is also supported by the VinUni-Illinois Smart Health Center at VinUniversity. }
\thanks{Phi V. Nguyen, Long Q. Tran are with the Institute for Artificial Intelligent, University of Engineering and Technology, Vietnam National University, Vietnam.}
\thanks{Hieu H. Pham are with the College of Engineering \& Computer Science, VinUni-Illinois Smart Health Center, VinUniversity, Vietnam and the Coordinated Science Laboratory, University of Illinois Urbana-Champaign, USA.}
\thanks{Copyright \textcopyright 2022 IEEE. Personal use of this material is permitted. However, permission to use this material for any other purposes must be obtained from the IEEE by sending a request to pubs-permissions@ieee.org.}}

\maketitle

\begin{abstract}
Convolutional neural networks (CNNs) have recently proven their excellent ability to segment 2D cardiac ultrasound images. However, the majority of attempts to perform full-sequence segmentation of cardiac ultrasound videos either rely on models trained only on keyframe images or fail to maintain the topology over time. To address these issues, in this work, we consider segmentation of ultrasound video as a registration estimation problem and present a novel method for diffeomorphic image registration using neural ordinary differential equations (Neural ODE). In particular, we consider the registration field vector field between frames as a continuous trajectory ODE. The estimated registration field is then applied to the segmentation mask of the first frame to obtain a segment for the whole cardiac cycle. The proposed method, Echo-ODE, introduces several key improvements compared to the previous state-of-the-art. Firstly, by solving a continuous ODE, the proposed method achieves smoother segmentation, preserving the topology of segmentation maps over the whole sequence (Hausdorff distance: 3.7–4.4). Secondly, it maintains temporal consistency between frames without explicitly optimizing for temporal consistency attributes, achieving temporal consistency in 91\% of the videos in the dataset. Lastly, the proposed method is able to maintain the clinical accuracy of the segmentation maps (MAE of the LVEF: 2.7–3.1). The results show that our method surpasses the previous state-of-the-art in multiple aspects, demonstrating the importance of spatial-temporal data processing for the implementation of Neural ODEs in medical imaging applications. These findings open up new research directions for solving echocardiography segmentation tasks.
\end{abstract}


\section{Introduction}
\label{sec:introduction}
\IEEEPARstart{E}{chocardiography} is one of the most popular non-invasive procedures for identifying structural abnormalities of the heart. It is inexpensive, requires no incisions, and can be performed quickly, enabling cardiologists to diagnose heart disease or assess a patient's risk of future cardiac events \cite{cheitlin1997acc}. Despite its ease of use and cost-effectiveness, ultrasound is well known for its noisy scans and variable image quality, which can depend on factors such as the skill of the person operating the device \cite{varoquaux2022machine}, making it challenging for automated systems to utilize the imaging data for patient diagnosis.

However, with the development of deep neural networks, particularly convolutional neural networks (CNNs), outstanding results have been achieved on multiple cardiovascular assessments, such as left ventricle segmentation or ejection fraction \cite{leclerc_deep_2019,ge_pv-lvnet_2019,ouyang_echonet-dynamic_2019}. Despite obtaining acceptable results on these evaluations, these models are still hindered in accurately defining two crucial phases of the cardiac cycle, namely end-diastole (ED) and end-systole (ES), due to the inherent noise present in ultrasound images and the high cost associated with annotating datasets. This limitation adversely affects the algorithm's ability to accurately segment the whole cardiac cycle, which is crucial for evaluating other parameters, such as global longitudinal strain \cite{smiseth2016myocardial, cikes2016beyond}.

While current methods for segmenting the cardiac cycle in echocardiography sequences have been introduced in the literature \cite{ge_pv-lvnet_2019, wei_temporal-consistent_2020, painchaud2022echocardiography}, they have several major limitations. For instance, approaches presented in \cite{ge_pv-lvnet_2019, painchaud2022echocardiography} either process individual frames or regularize the segmentation maps generated by models, failing to consider the crucial temporal relationship between consecutive frames in a video sequence. In contrast, \cite{wei_temporal-consistent_2020} addresses the temporal relationship between consecutive frames and utilizes a displacement field to regularize the segmentation maps, but it does not ensure the consistency and topology of the segmentation maps across the entire cardiac cycle.

To address these limitations, we present in this paper a novel approach for segmenting the cardiac cycle in echocardiography sequences. This approach allows generating a displacement field and propagating the segmentation map using a Neural ODE \cite{chen2018neural}. The motivation behind using the Neural ODE for segmentation in echocardiography video is to generate a smooth and topology-preserving registration from an ODE and to take advantage of the ability to learn dynamic and complex data from deep neural networks. Furthermore, incorporating segmentation from a 2D CNN model enables the production of a registration that includes spatial features from a conventional 2D segmentation model while also benefiting from the improved temporal representation provided by the Neural ODE. Additionally, by incorporating clinical loss functions, such as the mean absolute error (MAE) of the left-ventricular ejection fraction (LVEF), into the Neural ODE model, this approach can produce estimations of clinical parameters that are as accurate as those produced by the conventional 2D segmentation model. Finally, learning to estimate registration for propagating segmentation maps represents the underlying motion of the left ventricle chamber, which is beneficial for other clinical diagnoses in echocardiography imaging.

\textbf{Our contributions}: We summarize our main contributions as follows:

\begin{itemize}
    \item We present a novel echocardiography video segmentation method that propagates the segmentation map through the registration field obtained from a Neural ODE, Echo-ODE. The proposed approach optimizes both spatial and temporal information to effectively perform echocardiography segmentation;
    
    \item We introduce collection of loss function and regularization techniques that not only enhance segmentation performance but also improve the approach's effectiveness in terms of anatomical, cardiac motion, topology preservation, and clinical aspects;
    
    \item The proposed approach serves as post-processing for a trained segmentation model and optimizes a single video sequence, thereby eliminating the need for any additional training data;

    \item Extensive experiments on the CAMUS dataset demonstrate that our approach achieves state-of-the-art performance while showing interesting properties, e.g., preserving desirable topology and registration.
\end{itemize}

The rest of this paper is organized as follows: Previous works related to echocardiography segmentation are reviewed and discussed in Section \ref{sec:previous}. Section \ref{sec:framework} explains the proposed network architecture that is able to perform the diffeomorphic registration for echocardiography segmentation. Section \ref{exps} describes the dataset used in this study and the experimental settings. Experimental results and discussions are presented in Section \ref{results}. We also conduct in this section various experiments to investigate the model's properties. Finally, Section \ref{sec:conclusion} concludes the paper with future research directions.

\vspace{-0.08cm}
\section{Previous Works}
\label{sec:previous}

This section reviews recent literature on full cardiac cycle segmentation, especially on left ventricle segmentation and the Neural ODE. Details are below.

\vspace{-0.08cm}
\subsection{Echocardiography Segmentation}
\label{sec:echo_segmentation}

\subsubsection{Keyframe Segmentation}

Because of the expense of collecting and accurately annotating echocardiography data, previous works only focused on performing echocardiography segmentation using ES and ED images~\cite{leclerc_deep_2019, ouyang_echonet-dynamic_2019}. The first notable data collected were CAMUS by Leclerc \etal~\cite{leclerc_deep_2019} in 2019, which has 500 patients and was annotated by expert cardiologists. Within the same year, Ouyang \etal~~\cite{ouyang_echonet-dynamic_2019} published EchoNet which was designed for segmenting the left ventricle in A4C view sequences and estimating the ejection fraction (EF).

With the availability of large datasets, many conventional segmentation architectures were being assessed. In the same CAMUS work, Leclerc \etal~\cite{leclerc_deep_2019} evaluated various U-Net architect~\cite{ronneberger2015u} in terms of segmentation of the left ventricle (LV), myocardium (MYO) and left atrium (LA) for the ED and ES frames of the cardiac cycle for apical 4 chamber (A4C) and apical 2 chamber (A2C) views. Ouyang \etal~\cite{ouyang_echonet-dynamic_2019} also provided the segmentation result of DeepLabv3 ~\cite{chen2018encoder} architecture to the A4C view sequence. Beside experimenting with universal segmentation architectures, many methods were also proposed to specifically tackle problems on echocardiography datasets. Veni \etal~\cite{veni2018echocardiography} used a shape-driven deformable model with features provided by deep learning networks to regulate the segmentation mask. Thomas \etal~\cite{thomas2022light} proposed a novel model based on Graph Convolutional Networks, which was used to detect keypoints from ultrasound images. The result was then used to segment the LV and estimate the LVEF. 

Despite their success, the limitations of the above segmentation methods for the LV were inherent in their sole consideration of the ED and ES phases, ignoring the valuable temporal information that is crucial for accurate estimation of the EF, as demonstrated in work by Wei \etal~\cite{wei_temporal-consistent_2020}.

\subsubsection{Spatial Temporal Segmentation}
Recent papers \cite{ge_pv-lvnet_2019,li_mv-ran_2020,wei_temporal-consistent_2020} proposed specialized frameworks to focus on the temporal aspect of echocardiography segmentation. PV-LVNet~\cite{ge_pv-lvnet_2019} used temporal information extracted from recurrent networks to locate and crop the LV across whole US sequences. MV-RAN~\cite{li_mv-ran_2020} used both temporal and multi-view information to perform segmentation of the LV with a convolutional LSTM to capture temporal information. Wei \etal~~\cite{wei_temporal-consistent_2020} introduced CLAS, a co-learning shape and appearance model for temporal-consistent heart segmentation of echocardiography sequences with sparsely labeled data. The framework learned the 2D+time apical long-axis cardiac shape such that the segmented sequences can benefit from temporal and spatial consistency. While relying solely on ED and ES annotations, CLAS~\cite{wei_temporal-consistent_2020} also predicted deformation fields and used this information to regularize segmentation during training instead of propagating the segmentation to the whole sequence.

Another approach is the refinement of segmentation masks. Painchaud \etal~\cite{painchaud2022echocardiography} utilized AR-VAE~\cite{pati2021attribute} to post-process the latent representation of segmentation for every frame in the cycle to achieve the desired temporal consistency. The methods helped achieve SOTA results in LV segmentation and interpretable embedding space. However, the method did not account for temporal features from ultrasound but only used  segmentation results from a pretrained model, missed crucial temporal information presented in ultrasound sequence.

We address in this work the limitations of previous approaches by introducing a novel method, Echo-ODE, for echocardiography segmentation that leverages Neural ODE to model the temporal feature between frames, resulting in smooth and accurate segmentation. While prior research has focused primarily on keyframe segmentation and neglected valuable temporal information from ultrasound sequences, the method exploits the unique capabilities of Neural ODE to effectively capture the dynamic motion of the LV throughout the entire cardiac cycle.

\vspace{-0.08cm}
\subsection{Deformable Registration Learning}
\label{sec:representation_learning}
In the field of medical imaging, deformable image registration is used to find a dense, non-linear correspondence transform that can smoothly change one image or video into another~\cite{6522524}. Deformable registration for echocardiography can be used to register 3D models to 3D points, quantify myocardial motion and strain from multiple 3D ultrasound sequences, and segment the LV in 3D echocardiography \cite{eick2021associations}.

With the development of deep learning algorithms, researchers in computer vision and medical imaging are turning to deep neural networks for solving problems in the field of deformable image registration \cite{6522524}. One of the primary deep learning architectures used for deformable registration tasks is VoxelMorph \cite{balakrishnan2019voxelmorph}, which used a convolutional neural network to automatically create an accurate deformation field between two medical scans. Adversarial similarity networks were also used to improve the quality of the registration field by calculating the appearance loss between a warped image and a fixed image. In echocardiography processing, deformable image registration is also used to estimate the flow of motion between ultrasound images. A notable work by Zhu \etal~\cite{zhu2021test} that used U-Net~\cite{ronneberger2015u} to measure blood flow between ultrasound images at test-time training.

Solving for diffeomorphisms by putting the registration problem into the form of an ordinary differential equation (ODE) whose solution is a smooth deformation field is a promising direction. Dalca \etal~\cite{dalca2018unsupervised} extended VoxelMorph by forming the registration field as a diffeomorphic field of an ODE. By constraining the transformations to diffeomorphisms, the model from Dalca \etal~\cite{dalca2018unsupervised} could ensure certain desirable properties when transforming the image, such as the preservation of topology. The ODE itself could later be approximated by a neural network, which is more flexible and tractable for large-scale problems \cite{chen2018neural}. Xu \etal~\cite{xu2021multi} introduced a multi-scale Neural ODE for estimating registration of multi-contrast 3D MR images. Joshi \etal~\cite{joshi2021diffeomorphic} utilized Lipschitz continuous ResBlocks to solve the ODE that governs diffeomorphic deformations. Despite the success of these methods on different modalities, as far as we know, there is no work that uses diffeomorphic deformations from Neural ODE to solve the problem of echocardiography segmentation.

Our proposed method is an attempt to overcome the limitations of existing SOTA techniques. Ultrasound images are inherently noisy, which makes solving the Neural ODE without regularization and guidance inadequate for ensuring smooth and accurate registration while preserving the topology of segmentation maps throughout the entire sequence. To address this issue, we employ a balance between temporal consistency and segmentation map accuracy by incorporating both temporal features from the ultrasound sequence and segmentation maps from a pretrained segmentation model. Moreover, the proposed method integrates a clinical loss function, which facilitates the preservation of clinical accuracy for relevant metrics.
\begin{figure*}[tp]
    \includegraphics[width=\textwidth]{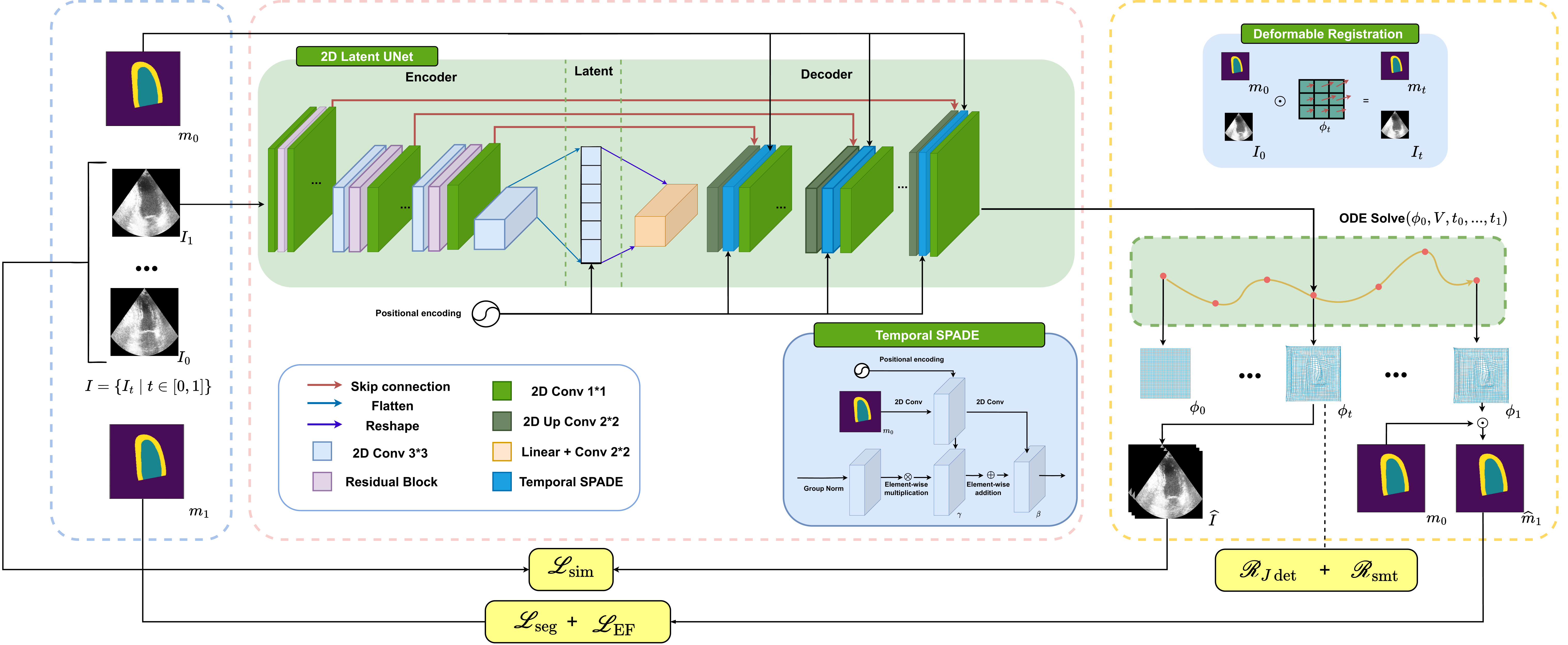}
    \caption{ An overview of Echo-ODE for segmenting echocardiography. Taking the echocardiography sequence $I$ and two segmentation map $m_0$ and $m_1$ as a starting point. At time step $t$, the the model encodes the ultrasound image, $I_t$, and decode it with an encoding of $t$ by a positional encoding in latent space. The model then generates an initial vector field by spatially adaptive normalizing the decoder features with positional embedding of $t$ and segmentation map $m_0$. The ODE solves for the deformation vector field $\phi_t$, used to generate a pseudo-image $\hat{I}_t$ and a pseudo-segmentation map $\hat{m}_t$. \vspace{-0.3cm}}
    \label{fig:method}
\end{figure*}

\section{Preliminaries}
\label{sec:preliminaries}

In this section, we provide an overview of the video segmentation and diffeomorphic registration methods that are the foundation of the proposed approach.

\subsection{Echocardiography Segmentation}

In echocardiography segmentation, given a sequence of ultrasound images from ED to ES, we have $I = \{ I_0, \dots , I_t, \dots , I_1 | t \in [0,1] \}$, where $I_0$ denotes the B-mode ultrasound image at the beginning of the cardiac cycle (ED), $I_1$ denotes the B-mode ultrasound image at the end of the cardiac cycle (ES), and $I_t$ denotes the intermediate image between ED and ES. Specifically, when $t \rightarrow 1$ the heart goes through systolic phase, undergoes contraction and ejects blood into the arteries. Each B-mode ultrasound image $I_t: \Omega \rightarrow \mathbb{R}$, where $\Omega$ is the image domain, has the segmentation map $\hat{m}_t : \Omega \rightarrow \mathbb{Z}$, in which each pixel $\hat{m}_t(p) = k$ corresponds to an object of interest, e.g., $k=1$ represents the LV region.

\subsection{Deformation Registration}
% The deformation field can then be used to transfer a smooth segmentation map from one image to another. Here, we assume that the images $I_t$  has segmentation mappings $m_t$. We can then use deformable registration to get segmentation map $m_{t+\Delta t}$ of the image $I_{t+\Delta t}$ : $m_{t+\Delta t}(p) = m_t(\phi_t(p))$.

Deformable registration aims to find an optimal transformation function that maps every point in the source image to a corresponding point in the target image, under some constraints of smoothness and preservation of topological properties. The transfer of segmentation maps is then accomplished by applying this deformation field to the source segmentation, effectively repositioning and reshaping the segmented structures to align with the corresponding structures in the target image. 

The deformable registration in our study could be formulated as follows. Given two B-mode ultrasound images at two time frames, beginning of cardiac cycle and arbitrary time, 0 and $t$, represented as $I_0, I_t$ . Let each pixel location be represented by $p \in \Omega \subseteq \mathbb{R}^2$. The goal of deformation registration is to find a transform $\phi_t : \Omega \rightarrow \Omega$ that minimizes a loss function $\mathcal{L}(I_t, \Phi_t(I_0))$, where $\Phi_t(I_0)(p) = I_0( \phi_t(p))$. Here $\mathcal{L}$ measures the dissimilarity between the image $I_{0}$ transformed by $\phi_t$ and the image $I_t$ under some regularization constrains $\mathcal{R}$. Given the segmentation map $m_0$ of $I_0$, our goal is to apply the previously computed transformation $\phi_t$ to transfer the segmentation map from the beginning of the cardiac cycle to an arbitrary time $t$. This process results in a new segmentation map $\hat{m_t}$ that aligns with the image $I_t$. 
% Because the transformation $\phi_t(p)$ may map a pixel to a non-integer location in the segmentation map $m_0$, the mapping is accomplished by applying the deformation field to the segmentation map using a suitable interpolation scheme, such as nearest-neighbor: $m_t(p) = \text{Interp}(m_0, \phi_t(p)), \forall p \in \Omega$.
%In other words, each pixel in the segmentation map $m_0$ is repositioned and potentially reshaped according to the transformation $\phi_t$ and the chosen interpolation method, ensuring that the transferred segmentation $m_t$ aligns with the anatomical structures in $I_t$. This allows for consistent segmentation across different time frames of the cardiac cycle.

% Previous works \cite{leclerc_deep_2019, wei_temporal-consistent_2020} treated the sequence as having a discrete and fixed length, $|S|$, thus constraining the smoothness of the sequence by the length of the sequence that the model processes. By formulating the registration field as a continuous registration field at generation time, we can create a sequence with an infinitely small time interval $\Delta t \to 0$ while estimating the registration field. This provides us with the flexibility to refine the smoothness of the sequence without requiring sophisticated post-processing methods and assumptions.

\subsection{ Diffeomorphic Registration Using Neural ODE}

To ensure the smoothness and topology preservation properties of the deformation, we define the deformation field as an ODE. Following the works of Rousseau \textit{et al.}\cite{rousseau2020residual} and Ashburner \cite{ashburner2007fast}, we parameterize the deformation registration $\phi_t$ by the ODE:

\begin{equation}
\frac{\mathrm{d} \phi_t(p)}{\mathrm{d} t}=V_{\theta}(p,t),
\label{eq:diff}
\end{equation}
where $\phi_t$ is the diffeomorphism at time $t$, and $V_{\theta}$ is the stationary velocity vector field parameterized by $\theta$, which is computed by a 2D convolutional neural network. By integrating the velocity field from time 0 to $t$, we obtain the deformation at any time $t$ ($\phi_t$), starting with the identity $\phi_0(p) = p $:

\begin{equation}
\phi_t(p) = \phi_0(p) + \int_{0}^{t} {V_{\theta}(\phi_s(p),s)} ds.
\label{eq:phit}
\end{equation}

In group theory, integrating this ODE to compute $\phi_t$ is analogous to computing the exponential map of the velocity field $V_\theta$ (which can be considered an element of a Lie algebra associated with a Lie group \cite{iserles2000lie}). The inverse transform $\phi_t^{-1}$ is obtained by integrating the negative of the velocity field.

% \begin{equation}
% \phi_t^{-1}(p) = p - \int_{0}^{t} {V_{\theta}(\phi_s^{-1}(p),s)} ds.
% \end{equation}
% \textcolor{red}{HP: a 2D convolutional neural network $f(\theta)$ ?} \pn{the $f(\theta)$ is about the output of the ODE model, we just refer to the parameter of the model only.}

\section{Proposed Method}
\label{sec:framework}

Our method focuses on echocardiography segmentation, where we employ diffeomorphic registration as a tool to improve the segmentation process. The method measures a diffeomorphic registration between first frame and arbitrary frame and uses the registration to propagate the initial segmentation map throughout the entire sequence. An overview of our method is visually represented in Fig.~\ref{fig:method}.

% \textcolor{red}{- People do not use A and B to denote images. You have to find another way to do that better. Here are some examples $\mathcal{A}$ $\mathcal{B}$ $\mathcal{X}$ $\mathcal{Y}$ $\mathcal{I}_t}$ 
% }

% \pn{ I think using mathbf to denote an image tensor is the norm, eg. \url{https://arxiv.org/pdf/2211.01600.pdf}, mathcal as I recall, is used to denote set. Papers from TMI actually use no format to denote image \$I\$ / $I$, eg. \url{https://arxiv.org/abs/2202.12498}, what do you think? }

\subsection{Learning registration field}
The deformation vector field $\phi_t$ must satisfy three criteria: (1) the deformation should be smooth; (2) the deformation vector field should represent the motion present in the ultrasound sequence; and (3) the topology of segmentation must be preserved after individual transformations, while the semantic information of the target segmentation, as learned by pre-trained segmentation models, must be accurately reflected in the applied transformations. Our method is illustrated in Fig. \ref{fig:method} and detailed in Algorithm \ref{algo:method}. We proposed a auto encoder model based on UNet \cite{ronneberger2015u} to effectively estimate the velocity field $V_\theta(p,t)$. The input for this model is the image $I_t$, segmentation map of the first frame $m_0$, and timestep $t$. The objective of the model is to estimate the registration field $V$ for ODE solver. 

\textbf{Encoder-Decoder}:  At the encoder module, the model encodes the image into a latent vector $v$ as a conventional auto-encoder model. At the decoder module, we augment the latent vector $v$ with cosine positional encoding for timestep $t$, as described in \cite{vaswani2017attention}, in order to maintain the order of the frame sequence, ensuring that the temporal relationships between frames are preserved. We also normalize decoder features to guide the flow field to follow LV, as described in the following section.

\begin{center}
\begin{algorithm} %should have [H] otherwise does not plot anything
\caption{Learning diffeomorphism $\phi: I_0 \rightarrow I_1$}
\label{algo:method}
\begin{algorithmic}[1]
    \footnotesize
    \Input Sequence images $I = \{ I_t\}_{t\in[0,1]}$, 2D model's segmentation maps $m = \{m_0, m_1\}$
    \Output{Registration fields $\phi = \{\phi_t\}_{t\in[0,1]}$ $\implies$ full cardiac segmentation $\hat{m}=\{\hat{m}_t\}_{t\in[0,1]}$}
    \algrule
    \Require  parameterized by neural networks:  $V_\theta$ in \eqref{eq:diff}.
    \algrule
    \While{V has not converged}
        \State Set  $\phi_0$ as identical transformation (vector field). 
            \For{$i=0,\ldots, 1$}
            \State Using Neural ODE($V_\theta,\phi_0)$ find a solution $\phi_t$.
            \State Given $\phi_t$, warp $I_0$ to $\hat{I}_t$
        \EndFor
        \State Given $\phi_1$, warp $m_0$ to $\hat{m}_1$
        \State update $V_\theta$ by minimizing {\it ODE loss}:
         $$ \mathcal{L}_{sim}(I, \hat{I}) + \mathcal{L}_{seg}(m_1, \hat{m_1}) + \mathcal{L}_{EF} +\mathcal{R}_{Jdet}(\phi) + \mathcal{R}_{smt}(\phi)$$
    \EndWhile
\end{algorithmic}
\end{algorithm}
\end{center}

\textbf{Temporal Spatial Adaptive Normalization}: Due to the fact that motion change between frames happens around the edge of anatomical structure and the correctly segments only happen on keyframe images, we extend SPADE~\cite{park2019SPADE} toward spatially-adaptive and time-dependent normalization by adding positional encoding of $t$, so the model can dynamically be guided by segmentation maps in the early part of the cardiac cycle, and use features from ultrasound later on. Specifically, in order to estimate the registration $V$, for each decoder feature map $f_n \subseteq \mathbb{R} ^ {H_n \times W_n \times C_n}$, we use the segmentation maps of the first frame $m_0$ and position encoding of timestep $t$ as inputs to normalize feature map $f_n$. We estimate two normalization parameters $\gamma_n, \beta_n \subseteq \mathbb{R} ^ {H_n \times W_n \times C_n}$. The normalized feature is given by:
$$ f_n = \gamma_n \otimes f_n \oplus \beta_n, $$
where $\otimes$ is element-wise multiplication and $\oplus$ is element-wise addition. This procedure is applied to each frame of the cardiac cycle, and the results are integrated by ODE solver, resulting in the following sequence of diffeomorphic registrations: $\phi = \{ \phi_t | i = 0, ..., 1 \}$.
% Leveraging the aforementioned sequence of vector fields $\phi$, we produce pseudo-images $\hat{I} = {\hat{I}_0, ..., \hat{I}_1}$ and pseudo-segmentation maps $\hat{m} = {\hat{m}_0, ..., \hat{m}_1}$. The segmentation map, specifically the segmentation of the first frame, is supplied by another segmentation model (e.g., 2D UNet \cite{ronneberger2015u}, DeepLab v3 \cite{chen2017rethinking}). The training objective focuses on minimizing the discrepancy between the pseudo-images $\hat{I}$ and ground truth images $I$ while preserving semantic similarity between pseudo-segmentation maps and the pre-trained model's segmentation maps.

\subsection{Loss Function and Regularization}
Following the work of Wu \etal~\cite{wu2022nodeo}, we design Algorithm \ref{algo:method} in order to find the optimal parameter $\theta$ that is satisfied:
\begin{equation}
    \begin{array}{r}
    \theta = \operatorname*{argmin}_\theta(\mathcal{L}_{sim} + \mathcal{L}_{seg} + \mathcal{L}_{EF} +\mathcal{R}_{Jdet} + \mathcal{R}_{smt}),
    \end{array} 
\end{equation}

\textbf{Similarity loss} is used to minimize the discrepancy between the pseudo-images $\hat{I}$ and the ground truth $I$. Given the prevalence of speckle noise in ultrasound images, we apply Gaussian kernels $\mathcal{K}$ (for individual images in the sequence) to both pseudo and ground truth images to smooth out images. The mean square error for each pixel is employed to minimize the difference between pseudo images $\hat{I}$ and the ground truth $I$, which is calculated as follows:

\begin{equation}
\mathcal{L}_{sim} = \mathbb{E}_{t \sim\mathcal{U}(0,1)} (I_t - \hat{I}_t)^2.
\end{equation}

\textbf{Segmentation loss} To enable the registration process to concentrate on the segmentation problem, we propose a secondary segmentation loss between the propagated segmentation maps and the segmentation maps generated by pre-trained segmentation models on keyframe only. The rationale for incorporating this additional loss is to guide the registration estimation to focus on the keyframe, where pre-trained model performed well, instead of whole sequence. The loss is calculated using the Dice coefficient as follows:
\begin{equation}
\mathcal{L}_{seg} = 2\frac{|\hat{m}_1 \cap m_1|}{|\hat{m}_1| + |m_1|},
\end{equation}
where $\hat{m}_1$ is the segmentation map from the registration model and $m_1$ is the segmentation map from the segmentation model at the ES frame. 

\textbf{Ejection Fraction loss} The main purpose of doing segmentation is to measure clinical parameters such as EF, LV mass, and LV volume. Painchaud \etal~\cite{painchaud2022echocardiography} showed that forcing the temporal consistency between segmentation maps caused the degradation of EF estimation in comparison to the segmentation maps from the pretrained segmentation model. To mitigate this problem, we propose a clinical attribute loss to minimize the difference between the EF obtained from segmentation maps from the registration model and EF obtained from segmentation maps from the pretrained segmentation model. The EF loss is calculated by:
\begin{align*}
\mathcal{L}_{EF} &= \left|\frac{EDV - ESV}{EDV} - \frac{EDV_{\text{pre}} - ESV_{\text{pre}}}{EDV_{\text{pre}}}\right| \label{eq:lef1}\\ 
&= 1 - \frac{\sum_{i} \sum_{j} \mathbbm{1}[\hat{m}_{1}(i, j) = \text{LV}]}{\sum_{i} \sum_{j} \mathbbm{1}[m_{1}(i, j) = \text{LV}] + \epsilon},
\end{align*}
where $EDV$ and $ESV$ are the end-diastolic volumes and end-systolic volumes from the registration model, and $EDV_{pre}$ and $ESV_{\text{pre}}$ are the end-diastolic volumes and end-systolic volumes from the pretrained segmentation model. Since only a 4 chamber view is available for EF estimation and $EDV = ESV_{\text{pred}}$, the volume is then approximated by the area (in pixels) of the LV segmentation map. Beside loss functions, we further add regularization terms to make sure the diffeomorphism vector field achieves smoothness and topology-preserving properties. 

\textbf{Topology preserving regularizer} The first regularization term is to ensure the Jacobian determinant of each deformable vector field $\phi_t$ is non-negative. The Jacobian determinant is positively making vector field $\phi_t$ create no fold during transformation, thus preserving the topology of the segmentation map after transformation by enforcing local orientation consistency of each pixel concerning neighbor pixels. Similar to NODEO~\cite{wu2022nodeo},  we minimize the negative elements of $\phi_t$ by apply ReLU activation function and use L2 norm to avoid sparsity of L1 norm, the regularization $\mathcal{R}_{Jdet}$ is given by:

\begin{equation}
 \mathcal{R}_{Jdet}=\mathbb{E}_{t \sim \mathcal{U}(0,1)}\left\|\textit{ReLU}\left(-\left(\left|\mathcal{D}_{\phi_t}(\mathrm{p})\right|\right)\right)\right\|_2^2, 
\end{equation}

\textbf{Smooth regularizer} The second term of regularization, $\mathcal{L}_{smt}$, serves to promote the smoothness of the transformation. This is achieved by minimizing the spatial gradients of the transformed image, thereby ensuring a seamless transition between adjacent pixels:
 
\begin{equation}
\mathcal{R}_{smt}=\mathbb{E}_{t,p}\left(\left\|\nabla_\phi(\mathrm{p})\right\|_2^2\right).
\end{equation}

\section{EXPERIMENTS}
\label{exps}

\subsection{Datasets and experimental settings}
\label{sec:setup}  

  Due to the lack of echocardiography datasets contains segmentation maps annotated for a full cardiac cycle. We evaluate the effectiveness of the proposed method for echocardiography segmentation tasks using only the CAMUS dataset \cite{leclerc_deep_2019,painchaud2022echocardiography}. The first version of the CAMUS dataset provides annotations for the ED and ES frames for each cardiac cycle. These frames were chosen to represent the peak of systole and diastole, respectively. The annotations in this version of the dataset encompass the locations of the left ventricle endocardium ($LV_{endo}$), myocardium ($LV_{epi}$), and left atrium ($LA$) boundaries. Designed for training segmentation models, the two-frame dataset comprises a total of 450 videos, each capturing a complete cardiac cycle (from ED to ES and from ES to ED). The dataset was subsequently partitioned into training and validation sets at a 90\% to 10\% ratio. All single-frame segmentation models employed to guide the ODE model were trained using this dataset.

The second version of the CAMUS dataset \cite{painchaud2022echocardiography} features full-cycle annotations for echocardiography videos, with an emphasis on temporal segmentation of the left ventricle. In particular, this dataset includes annotations for the endocardium ($LV_{endo}$) and myocardium ($LV_{epi}$) of the left ventricle for every frame from ED to ES. This version of the CAMUS dataset is utilized to assess the performance of segmentation methods developed using the two-frame dataset.
\subsection{Pretrained Segmentation Model}
\label{sec:segmentation-model}

We employed three distinct architectures to guide the learning of the registration field, namely DeepLabv3~\cite{chen2018encoder}, U-Net\cite{ronneberger2015u}, and ENet~\cite{paszke2016enet}. DeepLabv3~\cite{chen2018encoder} is a fully convolutional neural network (FCN) that utilizes atrous convolutions and a multi-scale feature fusion strategy to enhance semantic segmentation performance. U-Net~\cite{ronneberger2015u} is an FCN with a symmetric architecture that employs skip connections to combine information from various levels of resolution. ENet~\cite{paszke2016enet} is a lightweight network that leverages an encoder-decoder architecture to strike a balance between accuracy and computational cost.

\subsection{Implementation Details and Training Procedure}
\label{sec:training}
The training process for the registration field was conducted in two stages. The first stage involved training the segmentation models using the two-frame dataset, aiming to generate accurate segmentation maps of the LV. The segmentation maps created in the first stage served as a guide for estimating the registration field in the second stage on the full-cycle dataset.

In the initial stage, we trained the segmentation models using the two-frame dataset. All segmentation models were trained with the Dice loss function, a commonly used loss function for segmentation tasks. The parameters of the segmentation models were optimized using the Adam optimizer with a learning rate of 0.01 and a batch size of 8. Training was terminated when the validation loss ceased to improve. The segmentation models were trained for 100 epochs, utilizing a total of 720 training samples and 450 validation samples. The segmentation models were also evaluated using the Dice similarity coefficient (DSC) metric.

For the second stage, we estimated the registration field for each sample in the full-cycle dataset. This means that each model implicitly represents only a single echocardiography sequence. We used U-Net model architecture with an encoder consisting of four convolutional layers with channel sizes [16, 32, 32, 32] and a decoder comprising seven convolutional layers with channel sizes [32, 32, 32, 32, 32, 16, 16]. Each sample was optimized for 1000 steps. We used Euler's method with a step size of 0.1 to solve the ODE. The parameters of the ODE model were optimized using the Adam optimizer with a learning rate of 0.01.

\begin{table*}[tp]
  \centering
  \caption{Performance of the methods on temporal consistency, anatomical accuracy, and clinical metric. The Hausdorff distance (HD), mean absolute error of ejection fraction (EF) are reported in terms of the mean and the standard deviation across 98 test sequences, we also report mean Dice score and omit the standard deviation since we notice no significant difference between methods. The temporal error (Temp. Error) is the number of sequences that have at least one temporal inconsistency. The mDice is the mean Dice score of the LV myocardium and LV epicardium. The Dice LV$_{epi}$ and Dice LV$_{myo}$ are the Dice scores of the LV epicardium and LV myocardium, respectively. The best results are highlighted in bold.}
  \smallskip
  \begin{tabular*}{0.8\textwidth}
  {@{} @{\extracolsep{\fill}} l l cc ccc c @{}}
  \toprule
  \multirow{2}{*}{Methods}         &            & \multicolumn{2}{c}{Temporal}              & \multicolumn{3}{c}{Anatomical}                        & Clinical        \\
  \cmidrule(lr){3-4} \cmidrule(lr){5-7} \cmidrule(lr){8-8} 
                                   &            & \mcc{HD}      & \mcc{Temp. Error} & \mcc{mDice} & \mcc{Dice LV$_{epi}$}& \mcc{Dice LV$_{myo}$} & \mcc{EF}      \\
  \midrule
  CLAS~\cite{wei_temporal-consistent_2020}                             &           & 7.2 $\pm$ 5.2 &  98                &  0.911      & 0.898                &  0.925                & 6.0 $\pm$ 5.1 \\
  \midrule
  \multirow{3}{*}{Baseline}        & U-Net~\cite{ronneberger2015u}       & 4.6 $\pm$ 4.7 &  98                &  0.953      & 0.943                &  0.964                & 3.2 $\pm$ 2.6   \\
                                   & ENet~\cite{paszke2016enet}       & 4.6 $\pm$ 3.1 &  98                &  0.951      & 0.943                &  0.960                & 2.9 $\pm$ 2.5   \\
                                   & DeepLabv3~\cite{chen2018encoder} & 5.6 $\pm$ 4.7 &  98                &  0.945      & 0.935                &  0.955                & \bt{2.6 $\pm$ 2.3}   \\
  \midrule
  \multirow{3}{*}{Cardiac - VAE~\cite{painchaud2022echocardiography}}   & U-Net~\cite{ronneberger2015u}       & 4.2 $\pm$ 1.6 &  \textbf{0}                &  0.938      & 0.931                &  0.945                & 3.6 $\pm$ 3.3   \\
                                   & ENet~\cite{paszke2016enet}       & 4.3 $\pm$ 1.6 &  \textbf{1}                &  0.943      & 0.937                &  0.949                & 3.0 $\pm$ 2.4   \\
                                   & DeepLabv3~\cite{chen2018encoder} & 4.8 $\pm$ 1.5 &  \textbf{0}                &  0.943      & 0.937                &  0.949                & 3.3 $\pm$ 2.5   \\
  \midrule
  \multirow{3}{*}{Echo-ODE (ours)} & U-Net~\cite{ronneberger2015u}       & \textbf{3.7 $\pm$ 1.6} &  12                &  \textbf{0.957}      & \textbf{0.948}                &  \textbf{0.967}                & \textbf{3.1 $\pm$ 2.6} \\
                                   & ENet~\cite{paszke2016enet}       & \textbf{3.9 $\pm$ 1.6} &  14                &  \textbf{0.956}      & \textbf{0.948}                &  \textbf{0.963}                & \textbf{2.8 $\pm$ 2.5} \\
                                   & DeepLabv3~\cite{chen2018encoder} & \textbf{4.4 $\pm$ 1.8} &  9                &  \textbf{0.950}      & \textbf{0.940}                &  \textbf{0.959}                & 2.7 $\pm$ 2.3 \\
  \bottomrule                          
  \end{tabular*}
\label{tab:result}
\end{table*}

\subsection{Evaluation Metrics}
\label{sec:metric_exp_stt}

In this study, we employed a combination of three distinct metrics to evaluate the effectiveness of our proposed method: (1) temporal smoothness and consistency, (2) anatomical accuracy, and (3) clinical relevance. To assess segmentation quality, we report DSC for left ventricle endocardium ($LV_{endo}$), left ventricle myocardium ($LV_{epi}$), and mean DSC of those classes. In order to assess the smoothness of the segmentation maps, we calculate the average Hausdorff Distance (HD) between consecutive frames across the entire sequence. Although our model was not explicitly optimized to minimize temporal inconsistency, we also reported the seven inconsistency indicators proposed by \cite{painchaud2022echocardiography} to enable fair comparisons with other methods. The attributes used to characterize a 2D cardiac shape include: i) LV area, ii) LV width at the valves, iii) LV length from valves to apex, iv) orientation of the LV principal axis, v) myocardium (MYO) area, vi) horizontal center of mass of the epicardium (EPI), and vii) vertical center of mass of the EPI. To evaluate the clinical relevance of our approach, we used MAE of EF between segmentation maps generated by our method and ground truth, to assess the performance of our method in providing clinically relevant information. The temporal error indicates the number of sequences with at least one temporal attribute inconsistency.

\subsection{Baselines}
We conducted a comparative analysis of the proposed method against two categories of models: 1) popular 2D segmentation models, and 2) specialized temporal segmentation models for echocardiography. The popular 2D segmentation models that we compared against are U-Net~\cite{ronneberger2015u}, ENet~\cite{paszke2016enet}, and DeepLabv3~\cite{chen2018encoder}. In contrast, the specialized segmentation models we compared against are CLAS~\cite{wei_temporal-consistent_2020} and Cardiac AR VAE. CLAS~\cite{wei_temporal-consistent_2020} is a 2D segmentation model that is specifically designed for echocardiography segmentation. It employs an auxiliary registration loss to guide the segmentation model. On the other hand, Cardiac AR VAE is a temporal segmentation model that uses an Attribute-Regularized VAE (AR-VAE) to encode segmentation maps, utilizes a latent space that contains seven interpretable attributes to model the temporal dynamics of the left ventricle and nine residual attributes to encode additional information. The smooth segmentation maps is obtained by smooth out those attributes.

To evaluate the performance of the proposed and baseline methods, we trained the models on a two-frame dataset and evaluated them on a full-cycle dataset. The evaluation is conducted on 98 test sequences, and the results are reported in terms of mean and standard deviation over all test samples. We used default parameters from the original paper to train CLAS~\cite{wei_temporal-consistent_2020} and Cardiac AR VAE. All the baseline and proposed models were implemented using PyTorch and trained on a single NVIDIA Ampere A100 GPU.

\section{Results}
\label{results}

Table \ref{tab:result} presents the performance of five different methods on three evaluation metrics: temporal smoothness and consistency, anatomical accuracy, and clinical metric. In general, the Echo-ODE method outperforms the other methods in terms of temporal consistency, with the lowest Hausdorff distance of 3.7. Moreover, it achieves the highest mean Dice score of 0.967 for the LV myocardium and 0.957 for the LV epicardium, respectively. In terms of clinical metrics, our method achieved the lowest mean absolute error of ejection fraction of 2.7, and while not surpassing baseline models, our method still performs better than the current temporal SOTA. Overall, the Echo-ODE method shows promising results in accurately segmenting cardiac structures and ensuring temporal consistency in echocardiography sequences. The following sections discuss more details on different aspects of the results.

\subsection{Improvement over Anatomical Segmentation}

The results of methods on segmentation metrics are presented in Table \ref{tab:result}. While accurate segmentation is important, we believe that temporal smoothness and accuracy are equally critical factors that contribute to the effectiveness of our approach, and maintaining a consistent level of segmentation quality across different time points and images is essential for ensuring reliable diagnoses and treatments \cite{smiseth2016myocardial}. The proposed method achieved an average Dice coefficient at best of a mean Dice score of 0.957 for using U-Net~\cite{ronneberger2015u} as segmentation guidance, which is slightly better than the 2D segmentation models, which achieved a mean Dice score ranging from 0.945 to 0.953. Compared to models that utilized temporal information, our method outperforms Cardiac AR-VAE and CLAS~\cite{wei_temporal-consistent_2020} by a substantial margin. Furthermore, we also evaluated the results on different sub-regions of the $LV_{epi}$ and $LV_{myo}$, our model obtained an average Dice coefficient at best of 0.948 for $LV_{epi}$ and 0.967 for $LV_{myo}$, respectively, which further demonstrate the improvement in anatomical segmentation over baselines and previous SOTA using temporal information with only 0.943 for $LV_{epi}$ and 0.964 for $LV_{myo}$ at best. We also notice that the increase in segmentation accuracy is proportional to the accuracy of pre-trained segmentation models. For example, 2D U-Net~\cite{ronneberger2015u} achieves a better mean Dice score of 0.953 when compared to ENet~\cite{paszke2016enet} and Deeplabv3, and our method that utilizes U-Net~\cite{ronneberger2015u} also achieves better performance than ENet~\cite{paszke2016enet} and Deeplabv3. In a later section, we further investigate the impact of initial segmentation. 

\begin{figure*}[!ht]
    \centering
    \includegraphics[width=1\textwidth]{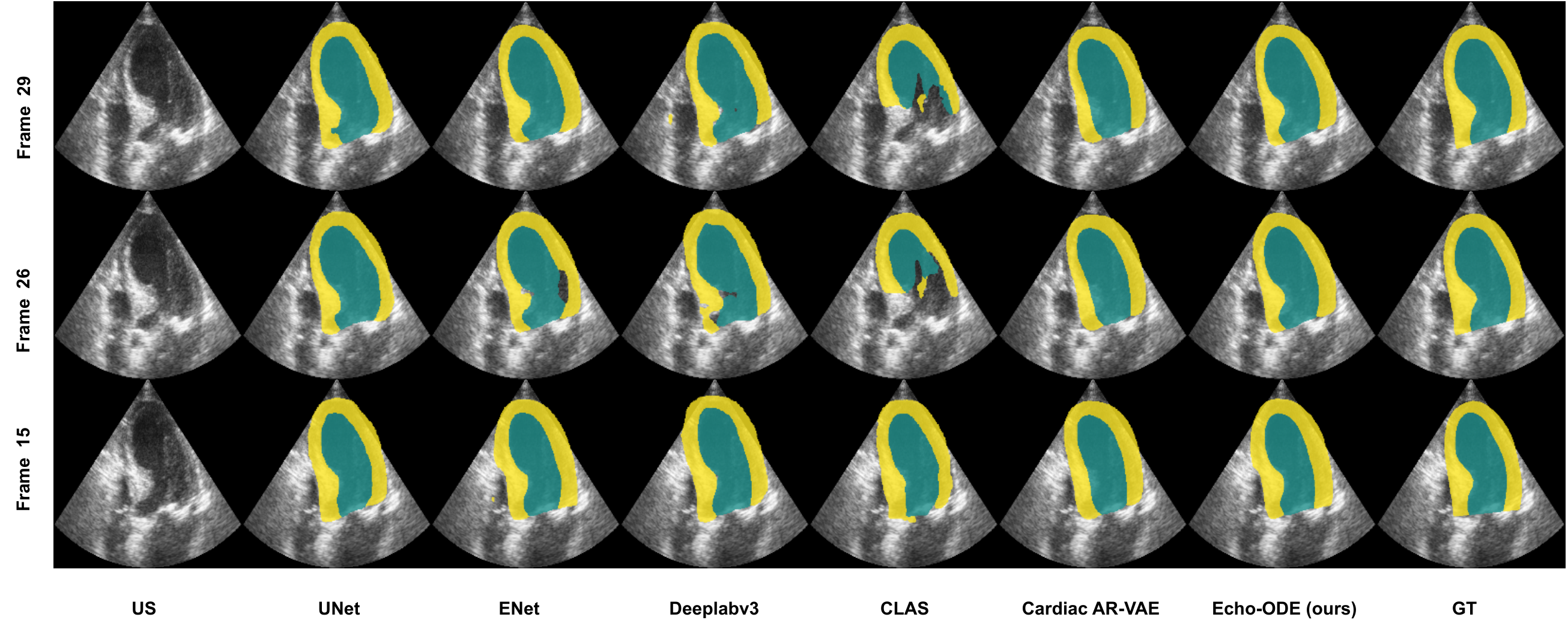}
    \vspace{-3mm}
        \caption{\textbf{Qualitative comparisons} of different methods on the typical frames from CAMUS datasets. Different color represents different semantic classes: $LV_{myo}$ (teal), $LV_{epi}$ (yellow) and background (purple). Both Cardiac AR-VAE and Echo-ODE use the same U-Net~\cite{ronneberger2015u} as input. More results can be found in the supplementary materials.} 
    \label{fig:results}
    \vspace{-1mm}
\end{figure*}
\subsection{Segmentation Smoothness and Temporal Consistency}

Assessing the temporal consistency of the segmentation maps is important for the clinical application of the segmentation models. The results are presented in the first column of Table \ref{tab:result}. The Echo-ODE method outperforms the other methods in terms of temporal smoothness with the lowest Hausdorff distance. Our method and Cardiac AR-VAE also produce segmentation maps with a lower standard deviation of Hausdorff distance of around 1.6, compared to 3.1-4.7 in baseline models, indicating that the temporal models produce consistent results. The results show that the proposed method achieved a mean Hausdorff distance of 3.7 to 4.4, which is lower than 4.2-4.8 for Cardiac AR-VAE and 7.2 for CLAS~\cite{wei_temporal-consistent_2020}, indicating the effectiveness of modeling the motion using neural ODE. Furthermore, we also detected 9 to 12 sequences that have at least one temporally inconsistent frame out of 98 sequences. While the temporal consistency of our method is not as good as Cardiac AR-VAE in the definition of Painchaud \etal~\cite{painchaud2022echocardiography}, it is still better than 2D segmentation methods and CLAS~\cite{wei_temporal-consistent_2020}. It is important to note that Cardiac VAE achieves these results by explicitly optimizing for the temporal consistency attribute, while optimizing the whole latent space of VAE does not yield such good results. The proposed method, on the other hand, does not require any assumptions from the data modality, which is better suited for a broader range of use cases.

\begin{figure}[tp]
    \begin{subfigure}[b]{\columnwidth}
        \centering
        \includegraphics[width=0.78\textwidth]{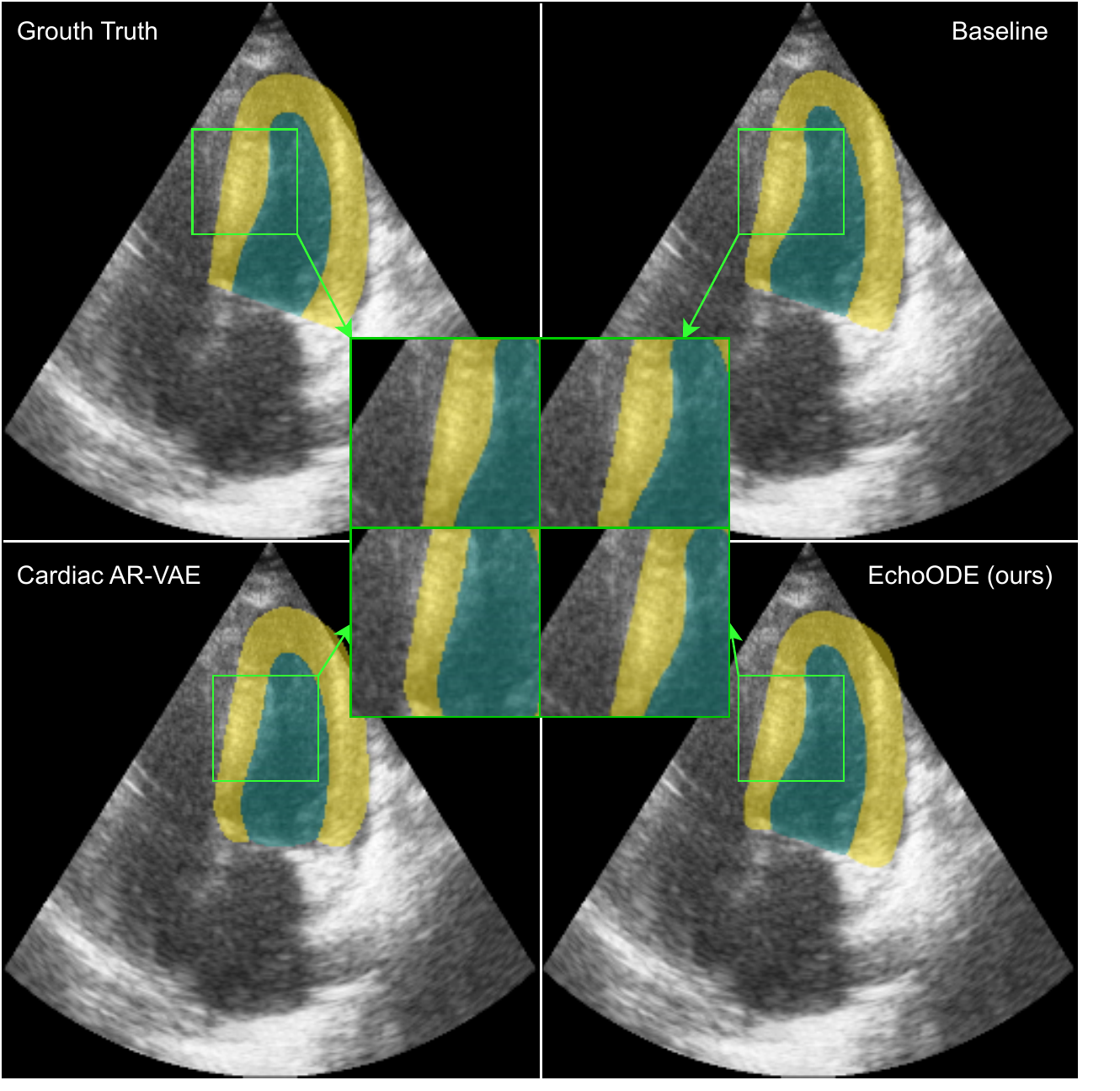}
        \caption{Highlight shape difference of segmentation maps, where the left half of endocardium of Cardiac AR-VAE appears to have different shape.}
    \end{subfigure}
    % \vfill
    \begin{subfigure}[b]{\columnwidth}
    \centering
        \includegraphics[width=0.8\textwidth]{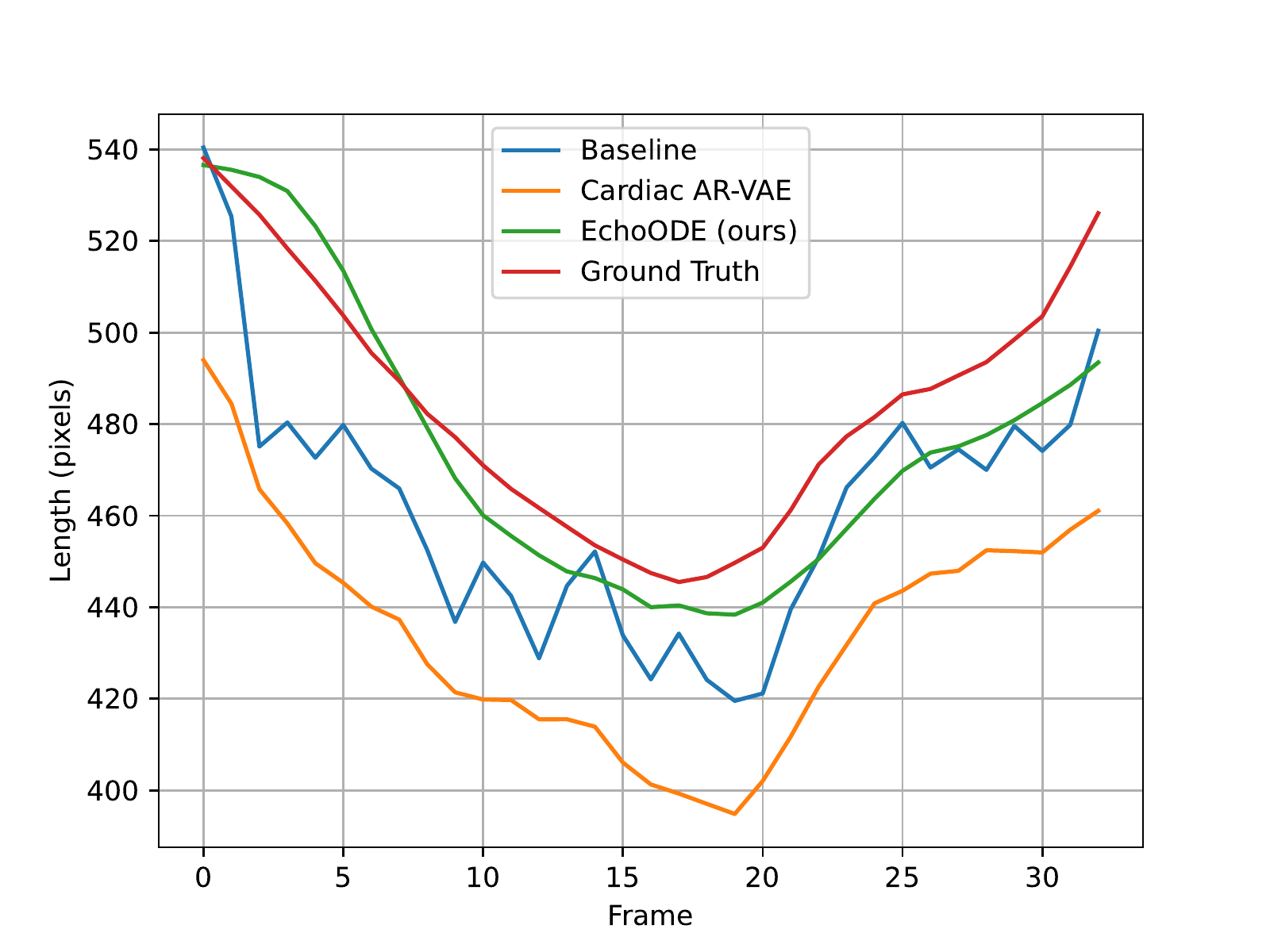}
        \caption{LV length from apex point to mitral valve of different models.}
    \end{subfigure}
    \caption{Visually and qualitative demonstration of different segmentation maps from different segmentation models.}
    \label{fig:shape}
\end{figure}

\subsection{Improvement over Clinical Metrics}

The data in Table \ref{tab:result} indicates that our proposed methods surpass CLAS~\cite{wei_temporal-consistent_2020} and Cardiac AR-VAE, which are specialized designed models, in terms of clinical metric. Specifically, our approach achieved a mean absolute error of 2.7 to 3.1 for EF, which is notably lower than the 0.6-0.7 range of Cardiac AR-VAE and 2.9-3.1 for CLAS~\cite{wei_temporal-consistent_2020}, demonstrating the accuracy of our method for estimating cardiac parameters. Additionally, our method does not experience a decline in clinical metrics when smoothing segmentation maps, which is a limitation of the Cardiac AR-VAE model. Nonetheless, we observed that our method does not yield significantly higher scores than 2D segmentation models in terms of clinical metrics, except for the U-Net~\cite{ronneberger2015u} and ENet~\cite{paszke2016enet} cases, which had a 0.1 higher score than the original 2D segmentation models. The other two models, ENet~\cite{paszke2016enet} and Deeplabv3~\cite{chen2018encoder} have slightly lower scores than 2D segmentation models. This is because our approach relies on segmentation maps and EF of baseline models to provide guidance, which are not as precise as ground truth segmentation maps from the outset. Future work will involve developing a better registration method that does not require segmentation maps for regularization.

\subsection{Shape correctness and localization}

When qualitatively evaluating the results of the previous SOTA, Cardiac AR-VAE, we observed a difference in the shape of the segmentation maps before and after being passed through the auto-encoder. This issue only occurred in the VAE model and was not present in the proposed method or other segmentation models. For instance, in Fig. \ref{fig:shape}(a), we can see that while the segmentation results from the original 2D segmentation models and the Echo-ODE produced shapes similar to the ground truth, the segmentation results from Cardiac AR-VAE tended to smooth out the contour of the segmentation maps. This caused the segmentation maps to diverge from the characteristic anatomical structure. To further illustrate the differences, we also plotted the length of the LV's apex and center of the mitral valve from the segmentation maps over time for different models in Fig.  \ref{fig:shape}(b). The figure indicates that Cardiac AR-VAE had a differing length of the LV, whereas our method had similar lengths and also had a smooth attribute over time. The cause of this issue may be due to the fact that the latent space of Cardiac AR-VAE encodes side information related to the shape of segmentation maps, and the smoothing action on the latent space over time may have inadvertently smoothed out the shape of the anatomical structure. On the other hand, other methods, including ours, do not require manipulation over the latent space and therefore do not suffer from this phenomenon.

\subsection{Impact of Initial Segmentation on Anatomical and Clinical Accuracy }

Our method's segmentation is obtained by propagating through the whole sequence from the first ED frame. Concerns may arise about the accuracy of segmentation maps from pre-trained models for the entire sequence. To assess the impact of the initial segmentation on the final results, we compared our method's segmentation using the initial frame from pre-trained segmentation models versus ground truth segmentation. Despite using the same registration field, we reported both anatomical and temporal metrics in Table \ref{tab:seg_gt}. The findings show that using ground truth as the initial segmentation map leads to better segmentation smoothness with a higher Hausdorff score of 0.1-0.6. However, the impact on anatomical accuracy and clinical metrics is minimal. The largest difference was observed with Deeplabv3~\cite{chen2018encoder}, where the Dice score was 0.08 lower when using a pre-trained segmentation map compared to the ground truth segmentation, while the MAE of EF remained the same. This suggests that the registration field is more accurate and smooth when close to the ground truth region, but using pre-trained segmentation maps as the initial segmentation is still effective for our method.

\begin{table}[!t]
    \caption{Performance of the proposed method on different segmentation model after replace initial segmentation map by the ground truth. all the results are reported in terms of mean across test set. The difference with using segmentation map of correspond segmentation model is reported in parenthesis. }
    \label{tab:seg_gt}
    \centering
    \smallskip
    \resizebox{0.5\textwidth}{!}{
    \begin{tabular}
            {@{} @{\extracolsep{\fill}} l c c c c c @{}}
            \toprule
            Segmentator & \mcc{HD}      & \mcc{mDice}      & \mcc{Dice LV$_{epi}$} & \mcc{Dice LV$_{myo}$} & \mcc{EF}      \\
            \midrule
            U-Net~\cite{ronneberger2015u}        & 3.6 (-0.1) $\pm$ 1.7    &  0.959(+0.02)      & 0.950(+0.02)                &  0.968(+0.01)                & 3.1(+0.00) $\pm$ 2.6 \\ 
            ENet~\cite{paszke2016enet}        & 3.6 (-0.3) $\pm$ 1.7    &  0.959(+0.03)      & 0.951(+0.03)                &  0.967(+0.04)                & 2.8(+0.00) $\pm$ 2.6 \\
            DeepLabv3~\cite{chen2018encoder}  & 3.8 (-0.6) $\pm$ 1.9    &  0.956(+0.06)      & 0.948(+0.08)                &  0.965(+0.04)                & 2.7(+0.00) $\pm$ 2.4 \\
            \bottomrule                          
    \end{tabular}
    }
\end{table}

\subsection{Topology Preserving and Registration}

This section presents the results of our method's ability to perform registration and preserve topology. 

\begin{figure}[tp]
\includegraphics[width=0.5\textwidth]{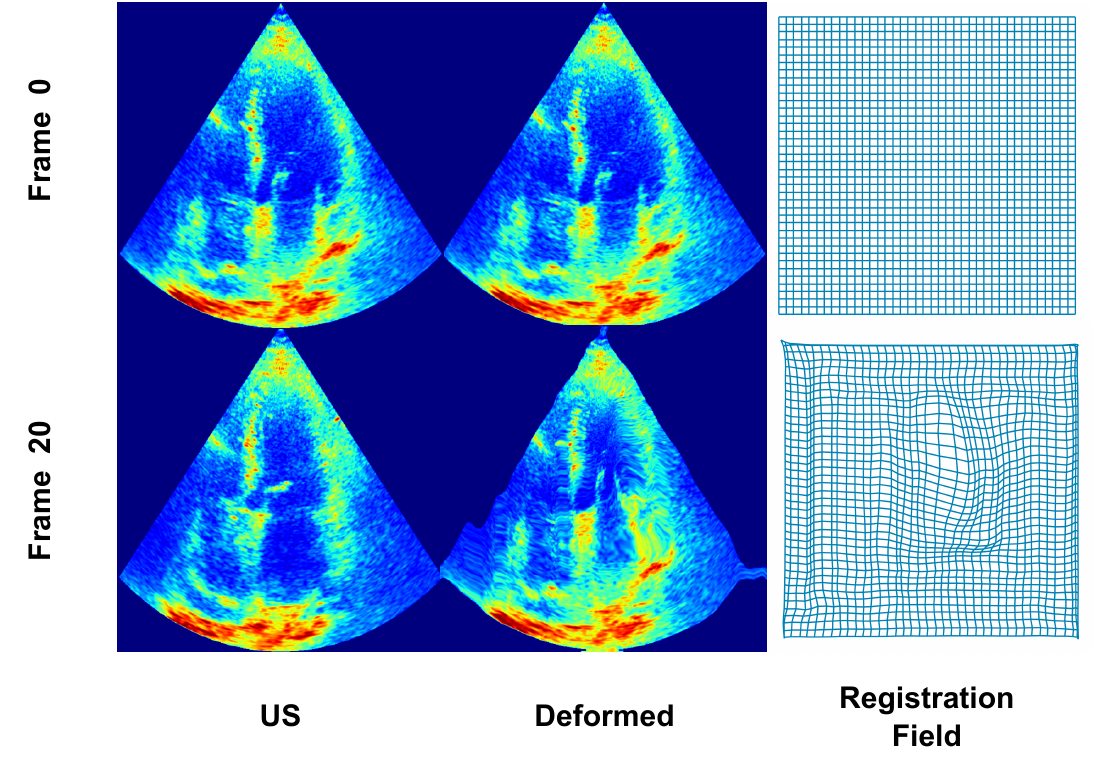}
\caption{\textbf{Topology preserving.} The registration results of our method (left) when applied to the first ultrasound frame of the sequence. The deformed grid (right) when applying the registration field (middle). The ultrasound frame is jet-colored for easier visualization of the speckle noise.}
\label{fig:temporal}
\end{figure}

Fig. \ref{fig:results} shows qualitative outcomes of different approaches on different sequence frames. U-Net~\cite{ronneberger2015u}, ENet~\cite{paszke2016enet}, and Deeplabv3 2D segmentation algorithms preserve the LV's shape but not its topology or smoothness. The CLAS~\cite{wei_temporal-consistent_2020} model maintains the shape correctness and smoothness of segmentation maps in early cardiac cycle but fails to preserve its topology later on. The Cardiac AR-VAE is more stable when the sequence is long but loses its LV shape after encoding and decoding the segmentation map. Our method, in the other hand, both preserves LV topology and segmentation map smoothness. Fig.  \ref{fig:temporal} demonstrates the registration field and how it transforms the image. For the first frame, the registration field is identity and the grid is not deformed, making the ultrasound identical to the original image. In the last frame, the ultrasound is deformed towards the real ultrasound image. The grid shows the topology preservation of the registration, where there is minimal folds or holes in the grid. It is important to note that the topology preservation of the registration field are obtained by minimizing the Jacobian determinant of the registration field. Therefore, there is no guarantee that the registration field will preserve the segmentation map's topology at all times, but it trying to preserve the topology of the registration field as much as possible. The method only ensure the topology of propagation, the correctness of segmentation maps would depend on the correctness of initial segmentation map. By using guidance from segmentation, we can see from the grid that the registration field follows the motion pattern of the LV, and the model attempts to register the ultrasound by following the speckle of the LV. This property allows us to use the registration field for other tasks, such as tracking the motion of the LV.

\subsection{Potential Limitations and Weaknesses}

\begin{figure}[tp]
\centering
\begin{subfigure}[b]{0.4\columnwidth}
\centering
\includegraphics[width=\textwidth]{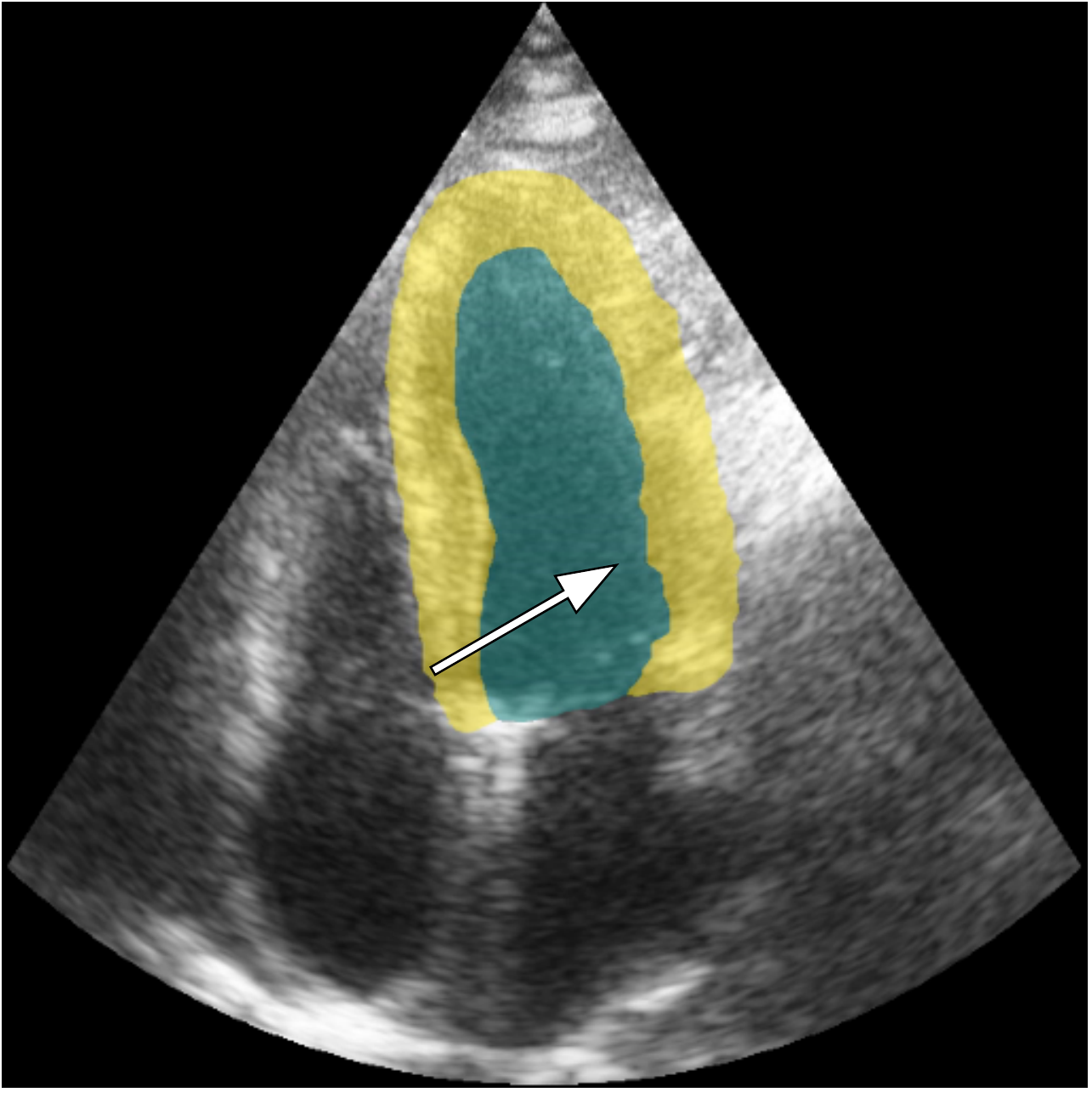}
\caption{An example of an incorrect LV segmentation map with a non-smooth contour due to noise.}
\end{subfigure}
\hspace{1em} 
\begin{subfigure}[b]{0.4\columnwidth}
\centering
\includegraphics[width=\textwidth]{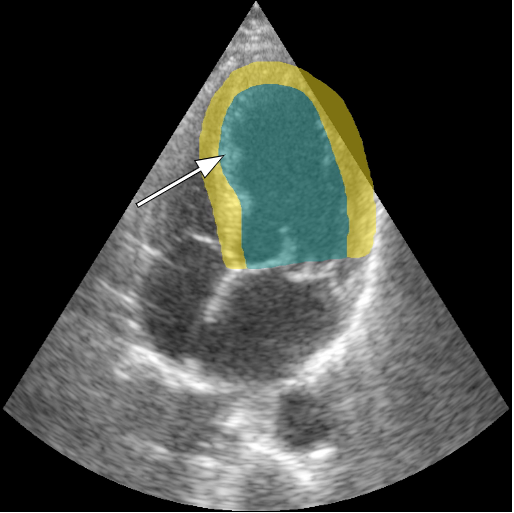}
\caption{In cases of rapid cardiac motion, the neural ODE solution may not accurately follow the heart's movements.}
\end{subfigure}
\caption{Illustration of the two most common errors observed in the results of our method.}
\label{fig:errors}
\end{figure}

Our method demonstrates promising results but has limitations, as visualized in Fig.  \ref{fig:errors}. Firstly, there's a trade-off between learning complex registration fields and contour continuity, with noise in fast-motion cases. This issue stems from pixel-level loss functions; future work will explore alternative loss functions for better contour capture. Secondly, the proposed method inconsistently localizes motion, as seen in Fig. \ref{fig:errors}(b). The global loss function minimizes mean absolute error, potentially leading to incorrect motion locations. Incorporating prior knowledge like optical flow could help, but the feasibility of deep learning remains uncertain. Lastly, the evaluation of registration and topology preservation warrants closer examination using quantitative and qualitative measures. We focused on video segmentation, which may not fully capture our method's performance in these aspects. Future work should involve comprehensive evaluation, the development of specific metrics, and the incorporation of diverse datasets for robust effectiveness assessment.

\section{Conclusion}
\label{sec:conclusion}

In this study, we introduced the Echo-ODE method, leveraging Neural ODE to enhance echocardiography segmentation. Our approach produces accurate and temporally consistent segmentation maps without explicit temporal optimization, outperforming state-of-the-art methods on the CAMUS dataset in terms of smoothness, anatomical accuracy, and clinical metrics. We also explored the impact of initial segmentation, finding minimal effects on accuracy and clinical metrics. Despite limitations in handling complex deformations and reliance on pre-trained models, the Echo-ODE approach shows promise for broader medical imaging analysis. Future directions include investigating advanced Neural ODE architectures, addressing limitations, and exploring uncertainty estimation techniques for more reliable segmentation maps. Additionally, exploring the method's applicability to other echocardiography processing and synthetic tasks could advance cardiac research.

% \appendices

% Appendixes, if needed, appear before the acknowledgment.

%\section*{Acknowledgment}

% The preferred spelling of the word ``acknowledgment'' in American English is 
% without an ``e'' after the ``g.'' Use the singular heading even if you have 
% many acknowledgments. Avoid expressions such as ``One of us (S.B.A.) would 
% like to thank $\ldots$ .'' Instead, write ``F. A. Author thanks $\ldots$ .'' In most 
% cases, sponsor and financial support acknowledgments are placed in the 
% unnumbered footnote on the first page, not here.

\bibliographystyle{IEEEtran}
\bibliography{citations}

\end{document}